\pdfoutput=1

\documentclass[11pt]{article}
\usepackage[]{acl}
\usepackage{times}
\usepackage{latexsym}
\usepackage[T1]{fontenc}
\usepackage[utf8]{inputenc}
\usepackage{microtype}
\usepackage{inconsolata}
\usepackage{graphicx}
\usepackage[utf8]{inputenc}
\usepackage{times}
\usepackage{xcolor}
\usepackage{graphicx}
\usepackage{amsmath,amsfonts,amssymb}
\usepackage{booktabs}
\usepackage{algorithmic}
\usepackage{algorithm}
\usepackage{array}
\usepackage{arydshln}
\usepackage{multirow}
\usepackage{url}
\usepackage{verbatim}
\usepackage{cite}
\usepackage{threeparttable}
\usepackage{balance}
\usepackage{colortbl}
\usepackage{makecell}
\usepackage{utfsym} 
\usepackage{CJK}
\usepackage{xspace}
\usepackage{bm} 
\usepackage{bbm}

\definecolor{Mygreen}{RGB}{113, 174, 72}
\definecolor{Myyellow}{RGB}{255, 205, 0}
\definecolor{Myred}{RGB}{192, 0, 0}

\makeatletter
\DeclareRobustCommand\onedot{\futurelet\@let@token\@onedot}
\def\@onedot{\ifx\@let@token.\else.\null\fi\xspace}
\def\eg{\emph{e.g}\onedot}
\def\ie{\emph{i.e}\onedot}

\def\vs{\emph{vs}\onedot}

\makeatother
\newcommand*{\email}[1]{\texttt{#1}}

\title{Enhancing Partially Relevant Video Retrieval with Robust Alignment Learning}

\author{Long~Zhang\textsuperscript{1}, Peipei~Song\textsuperscript{1}\footnotemark[1], Jianfeng~Dong\textsuperscript{3}, Kun~Li\textsuperscript{4} \and Xun~Yang\textsuperscript{1,2}\thanks{Corresponding author.} \\
  \textsuperscript{1}University of Science and Technology of China \\
  \textsuperscript{2}MoE Key Laboratory of Brain-inspired Intelligent Perception and Cognition, \\ University of Science and Technology of China
 \\
  \textsuperscript{3}Zhejiang Gongshang University
  \hspace{0.6em}
    \textsuperscript{4}ReLER, CCAI, Zhejiang University\\
  \email{dragonzhang@mail.ustc.edu.cn, beta.songpp@gmail.com} \\
  \email{dongjf24@gmail.com, kunli.hfut@gmail.com, xyang21@ustc.edu.cn} \\
}

\begin{document}
\maketitle
\begin{abstract}
Partially Relevant Video Retrieval (PRVR) aims to retrieve untrimmed videos partially relevant to a given query.  
The core challenge lies in learning robust query-video alignment against spurious semantic correlations arising from inherent data uncertainty: 1) query ambiguity, where the query incompletely characterizes the target video and often contains uninformative tokens, and 2) partial video relevance, where abundant query-irrelevant segments introduce contextual noise in cross-modal alignment. Existing methods often focus on enhancing multi-scale clip representations and retrieving the most relevant clip. However, the inherent data uncertainty in PRVR renders them vulnerable to distractor videos with spurious similarities, leading to suboptimal performance.
To fill this research gap, we propose Robust Alignment Learning (RAL) framework, which explicitly models the uncertainty in data.
Key innovations include: 1) we pioneer probabilistic modeling for PRVR by encoding videos and queries as multivariate Gaussian distributions. This not only quantifies data uncertainty but also enables proxy-level matching to capture the variability in cross-modal correspondences; 2) we consider the heterogeneous informativeness of query words and introduce learnable confidence gates to dynamically weight similarity.
As a plug-and-play solution, RAL can be seamlessly integrated into the existing architectures.
Extensive experiments across diverse retrieval backbones demonstrate its effectiveness.
\end{abstract}
\vspace{-0.4cm}

\begin{figure}[t]
    \centering
    \includegraphics[width=\columnwidth]{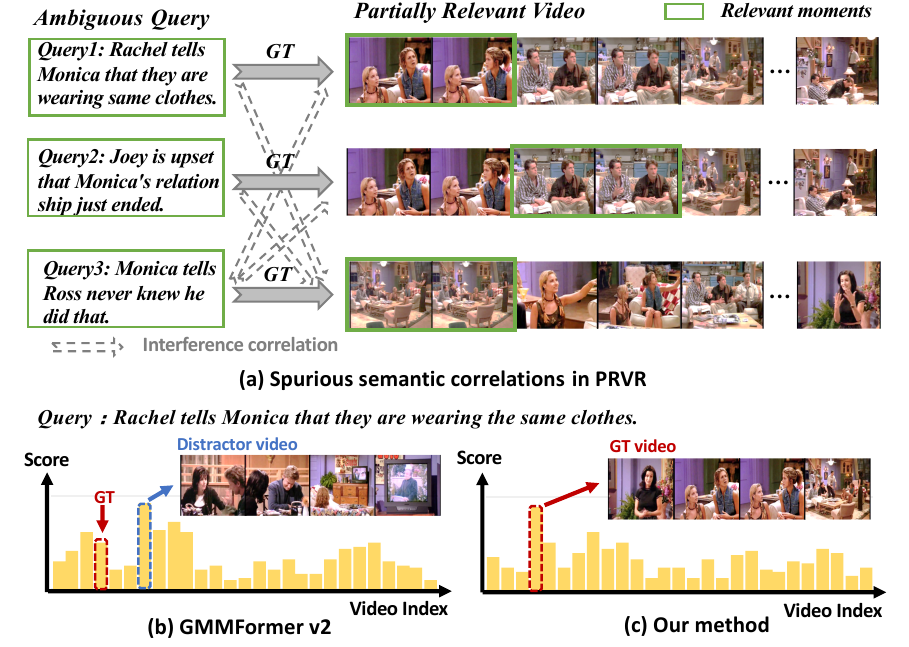}
    \vspace{-0.6cm}
    \caption{
    (a) Toy examples of spurious semantic correlations. (b--c) Retrieval scores of our method \vs GMMFormer v2. GMMFormer v2 fails to handle uncertainty, assigning the highest score to a distractor video. }
    \label{fig:query-type}
    \vspace{-0.6cm}
\end{figure}

\section{Introduction}
Text-to-Video Retrieval (T2VR) has been a long-standing challenge in vision and language research, allowing humans to associate textual concepts with video entities \citep{wang2025episodic, jin2023video, bogolin2022cross,yang2022video,yang2024robust}. However, the mainstream T2VR methods \citep{li2024momentdiff,wu2023cap4video,wang2023unified} assume that videos are pre-trimmed and text queries fully correspond to the videos \citep{dong2023dual}. In real-world scenarios, videos are often untrimmed and the given queries can be incomplete and ambiguous, describing only a portion of the target video. This realistic demand leads to the emergence of Partially Relevant Video Retrieval (PRVR) task \citep{wang2023gmmformer, dong2022partially}, which aims to find untrimmed videos that are only partially relevant to a given text query.

PRVR presents a fundamental challenge of spurious semantic correlations due to the query ambiguity and partial video relevance. As illustrated in Figure \ref{fig:query-type} (a), such spurious correlations manifest in two aspects: the query `` {Monica tells Ross never knew he did that}" relates to multiple video segments featuring similar actions in different contexts (\textit{query ambiguity}), while the target video contains diverse content described by multiple sentences (\textit{video partial relevance}). These factors make it difficult to establish a robust query-video alignment. 
Existing PRVR methods primarily attempt to mitigate query ambiguity by learning multi-scale clip representations, thereby maximizing the query-clip similarity within positive query-video pairs \citep{dong2023dual, wang2023gmmformer}. However, they implicitly assume a deterministic query-clip mapping and overlook the inherent data uncertainty in PRVR, thereby reducing inherently complex semantic mappings to deterministic pointwise alignments. Besides, without moment-level annotations, these methods struggle to learn optimal clip representations, leading to performance bottlenecks. Furthermore, they may be influenced by distractor videos with similar segments and provide incorrect retrieval results, as shown in Figure \ref{fig:query-type} (b).

To address the above issues, we propose Robust Alignment Learning (RAL), which explicitly models and utilizes uncertainty in data to enhance retrieval robustness. Our RAL builds upon the insight that PRVR should not be treated as point-wise query-clip feature alignment but rather as a probabilistic alignment problem that accounts for uncertainty. Inspired by probabilistic distributional representations \citep{jin2022expectation}, we model both video and query embeddings as Gaussian distributions, where the variance quantifies the inherent aleatoric uncertainty in each instance. {Based on the distributional representations}, we naturally construct Gaussian-based text and video proxies, which serve as multiple potential alignment candidates, enabling the model to capture diverse cross-modal relationships. 
Furthermore, most retrieval methods compute similarity scores by applying mean-pooling over words in the word-frame similarity matrix \citep{zhang2023multi, zhang2025multi}. We find this approach exacerbates retrieval bias as not all words contribute equally to retrieval, meaningless words (\eg, ``a'') can distort the similarity estimation. To address this, we introduce confidence-aware alignment that dynamically assigns confidence weights to query words.

\begin{figure*}[t]
    \centering
    \includegraphics[width=\textwidth]{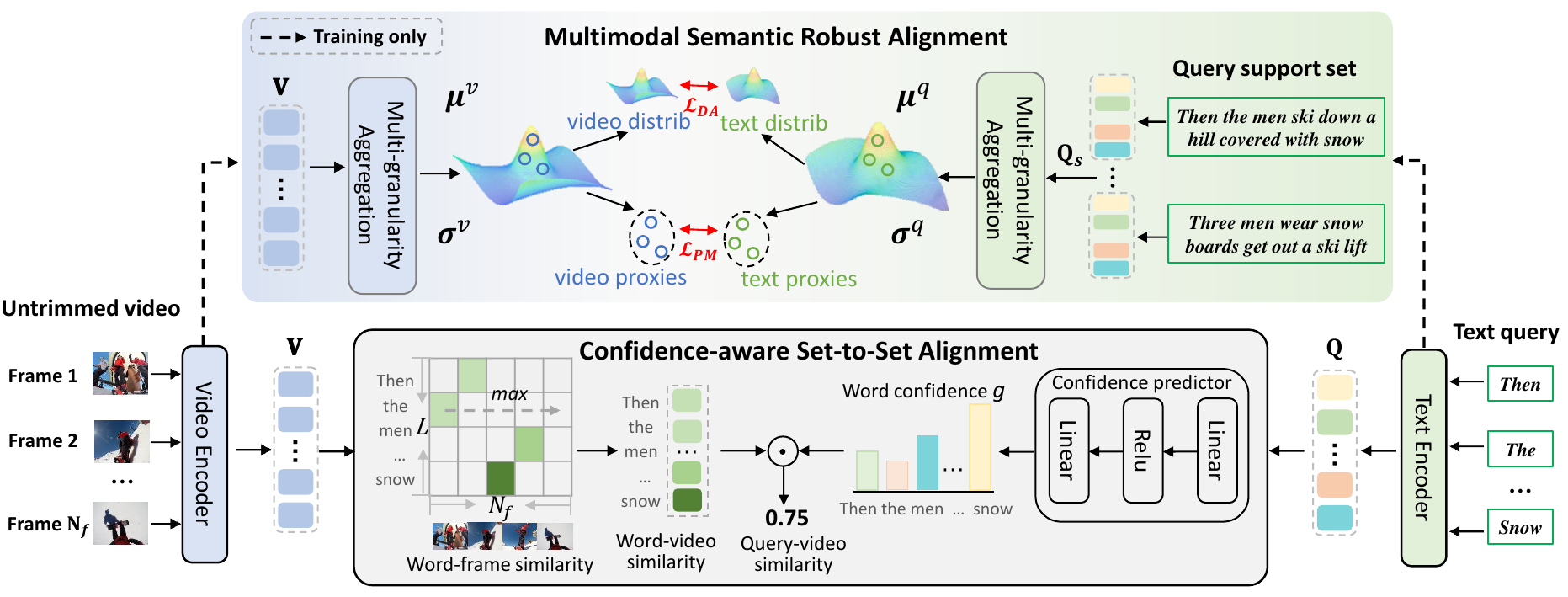}
    \vspace{-0.4cm}
    \caption{Overview of the proposed framework. It mainly consists of two components: (1) Multimodal Semantic Robust Alignment (MSRA) and (2) Confidence-aware Set-to-Set Alignment (CSA). Given an untrimmed video and a query, we first extract the frame features ${\bf V}$ and word features ${\bf Q}$ by video and text encoder, respectively. For MSRA, we collect a query support set containing all queries related to the video, obtaining its features ${\bf Q}_s$ with rich contexts. Then, we apply multi-granularity aggregation to obtain holistic semantics, and generate distributional representations parameterized by mean vector ${\bm \mu}$ and variance vector ${\bm \sigma}$. A proxy matching loss $\mathcal{L}_{\mathrm{PM}}$ and a distribution alignment loss $\mathcal{L}_{\mathrm{DA}}$ are used to unify the video and text domains. For CSA, we adopt a confidence predictor to assign confidence weights to each word, which is used to adjust the word-frame similarity matrix for video retrieval.
    }
    \label{fig:main}
    \vspace{-0.6cm}
\end{figure*}

As shown in Figure \ref{fig:main}, our RAL consists of two key components: (1) \textbf{Multimodal Semantic Robust Alignment (MSRA)} quantifies the semantic distribution in each modality by representing samples as multivariate Gaussian distributions. Given video and query embeddings, we first employ multi-granularity aggregation to obtain holistic semantics with sufficient contexts before estimating Gaussian parameters. 
Considering the incompleteness of the query relative to the video, we construct text distribution from a query support set that combines all video-related queries. 
Then, we conduct cross-modal learning with these distributional representations to joint video and text domains.  
To be specific, MSRA is optimized with two losses: a distribution alignment loss $\mathcal{L}_{\mathrm{DA}}$ enforcing probabilistic alignment between video and text distributions for robust cross-modal consistency, and a proxy matching loss $\mathcal{L}_{\mathrm{PM}}$ leveraging multiple alignment candidates to capture diverse semantic relationships.
(2) \textbf{Confidence-aware Set-to-Set Alignment (CSA)} is to enhance query-video matching by dynamically adjusting the contribution of each query word.
Instead of treating all words equally, CSA predicts a confidence score for each word and uses it to weight the word-frame similarity matrix. This effectively mitigates the influence of meaningless words and improves video retrieval.

Our contribution can be summarized as follows:
\vspace{-0.6cm}
\begin{itemize}
\setlength{\itemsep}{0pt}
\setlength{\parsep}{0pt}
\setlength{\parskip}{0pt}
    \item {We propose a novel robust alignment learning method for PRVR. It explicitly models and utilizes the data uncertainty and considers multiple potential matching relationships to enhance retrieval robustness.}
    \item We propose a confidence-aware dynamic weighting mechanism for query words, which effectively mitigates the matching noise brought by meaningless words, improving retrieval precision.
    \item{Extensive experiments on benchmark datasets (\ie, TVR \citep{lei2020tvr} and ActivityNet \citep{krishna2017dense}) demonstrate that our RAL significantly improves existing methods, achieving state-of-the-art results on PRVR. }
\end{itemize}

\section{Related Work}
\textbf{Partially Relevant Video Retrieval}
PRVR aims to retrieve untrimmed videos partially relevant to a given query. Compared to traditional T2VR, this task is more aligned with real-world application scenarios. Existing research \citep{dong2022partially,wang2023gmmformer,wang2024gmmformer,jiang2023progressive,nishimura2023large,dong2023dual,Song2025AMD,cho2025ambiguity,zhang2025multi} primarily tackled PRVR by constructing multi-scale clip representations. Specifically, MS-SL \citep{dong2022partially} applies sliding windows to form clip representations and performs similarity calculations at both clip and frame levels. GMMFormer \citep{wang2023gmmformer} uses multiple Gaussian windows to constrain inter-frame interactions, thereby implicitly generating multi-scale clip features. Its improved version, GMMFormer v2 \citep{wang2024gmmformer}, introduces a learnable feature fusion mechanism to aggregate multi-scale clips.  
Despite promising advancements, these methods suffer from performance bottlenecks due to ignoring the spurious semantic correlations caused by data uncertainty and simplifying the complex semantic alignment, which motivates our robust alignment learning method.

\textbf{Uncertainty in Multimodal Learning}
Uncertainty modeling has been widely explored in multimodal learning \citep{gao2024embracing}. HIB \citep{joon2019HBI} first introduces probabilistic embeddings to capture the uncertainty in image representations. Similar ideas have been applied to tasks such as sentiment analysis \citep{gao2024embracing} and instance segmentation \citep{zhang2021point}. In the field of cross-modal retrieval, {PCME \citep{chun2021probabilistic} pioneers the use of probabilistic embeddings to capture the uncertainty of visual concepts.} UATVR \citep{fang2023uatvr} further combines deterministic and probabilistic embeddings to explore optimal matching granularity in T2VR. T-MASS \citep{wang2024text} introduces a text-mass-based method, treating text embeddings as stochastic variables.  
\textit{However}, these methods are typically designed for trimmed videos and exhibit limited effectiveness in PRVR. 
Inspired by this, we propose robust alignment learning specifically designed for PRVR. 

\section{Method}
\subsection{Preliminaries}
\label{pipeline}
In this paper, we tackle the task of PRVR. 
Given a text query $q$ and a gallery of untrimmed videos $\mathcal{V}$, the goal of PRVR is to rank all videos $v\in \mathcal{V}$ so that the video partially corresponding to the text query $q$ is ranked as high as possible.
Existing methods primarily rely on multi-scale clip modeling to capture one-to-one correspondences between queries and untrimmed videos implicitly \citep{wang2023gmmformer, dong2023dual}. Here, we first review the common retrieval pipeline. 
For a query-video pair $(q,v)$, unimodal encoders extract frame features \({\bf V}\in \mathbb{R}^{N_f\times d}\) and word features \({\bf Q}\in \mathbb{R}^{L\times d}\), where $N_f$ and $L$ denote the number of frames and words, respectively. Both features are projected into $d$-dimension feature space for cross-modal retrieval. Then, the clip modeling module (\eg, sliding windows \citep{dong2022partially} and Gaussian windows \citep{ wang2023gmmformer}) is applied on $\textbf{V}$ to form the clip embeddings $\{{\bf c}_1,...,{\bf c}_{N_c}\}$. Meanwhile, attention pooling summarizes $\textbf{Q}$ into a sentence embedding ${\bf q}$. The final retrieval score $S(q,v)$ is computed as the maximum cosine similarity between sentence and clip embeddings:
\begin{equation}
\label{eq: old_score}
    S(q,v) = \max(\cos( \textbf{q}, \textbf{c}_1), \ldots, \cos( \textbf{q},\textbf{c}_{N_c})). \quad
\end{equation}
To enforce cross-modal alignment, existing methods typically optimize a combination of InfoNCE contrastive loss \(\mathcal{L}_{nce}\) \citep{miech2020end} and triplet ranking loss \(\mathcal{L}_{trip}\) \citep{dong2022partially}:
\begin{equation}
    \mathcal{L}_{base} = \lambda_1\mathcal{L}_{nce} + \lambda_2\mathcal{L}_{trip},
\end{equation}
where $\lambda_{1}$ and $\lambda_{2}$ are hyperparameters to balance the losses. $\mathcal{L}_{base}$ encourages high query-clip similarity $S(q,v)$ within positive query-video pairs while pushing apart negatives. 

\textbf{Motivation.} In other words, this common pipeline implicitly assumes a deterministic mapping between a query and a video clip. However, this assumption is problematic given the query ambiguity and partial video relevance, \ie, uncertainty in data. To address limitations, we are devoted to explicitly modeling the data uncertainty and leveraging it to augment the query and video representations, thereby improving the robustness of retrieval.

\subsection{Multimodal Semantic Robust Alignment}
Considering the query ambiguity and partial video relevance, we first propose an MSRA module to quantify the aleatoric uncertainty within both modalities. By modeling this uncertainty, we can better capture the variability in cross-modal correspondences and leverage it to strengthen cross-modal learning, enabling more robust representations of text and video.

\noindent\textbf{(1) Uncertainty Modeling.} 
According to uncertainty estimation theories \citep{chun2021probabilistic, gao2024embracing}, the aleatoric uncertainty can be predicted with deep learning models as the Gaussian variance. Inspired by this, we model the uncertainty in PRVR by treating feature representations as Gaussian distributions. Given the preliminary embeddings $\textbf{X}^{m}$ of input $m$ ($m\in\{q,v\}$), we estimate the mean vector ${\bm \mu}^{m}\in\mathbb{R}^d$ and variance vector ${\bm \sigma}^{m2}\in\mathbb{R}^d$ through two fully connected layers: 
\begin{equation}
\label{eq:mu}
        \boldsymbol{\mu}^m = {h_{\mu}^m}(g^m(\mathbf{X}^m)), \ \boldsymbol{\sigma}^m = {h_{\sigma}^m}(g^m(\mathbf{X}^m)), 
\end{equation}
where ${h_{\mu}^m}(\cdot)$ and ${h_{\sigma}^m}(\cdot)$ are the mean and variance estimators for input $m$, and $g^m(\cdot)$ serves as the feature aggregator. 
Furtherly, we define the probabilistic representation $\textbf{z}^{m}$ as a multivariate Gaussian distribution with \( d \) variable: 
\begin{equation}
\label{eq:uncert}
        p(\mathbf{z}^m|\mathbf{X}^m) \sim \mathcal{N}(\boldsymbol{\mu}^m, \boldsymbol{\sigma}^{m2}\mathbf{I}),
\end{equation}
where $\mathbf{I}$ is the identity matrix. 
{The uncertainty-aware representation $p(\mathbf{z}^m|\mathbf{X}^m)$ allows the model to capture variability in semantic alignment.}

\textit{\textbf{Query Support Set:} }
For text modality, a single query provides an incomplete description of the video, limiting the reliability of its probabilistic representation. To this end, we replace the standalone query embedding $\textbf{Q}$ with an enriched query support set embedding $\textbf{Q}_s$, for better textual uncertainty modeling. Specifically, for each video $v$, we construct a query support set $\mathcal{D}^v$ by aggregating all associated queries $q_n$. The $\textbf{Q}_s$ is obtained by concatenating the embeddings of all $q_n$ in $\mathcal{D}^v$: 
\begin{equation}
    \mathbf{Q}_s = ||_{q_n\in\mathcal{D}^v}(\mathbf{Q}_n),\ \mathcal{D}^v=\{q_n|q_n\Leftrightarrow v\},
\end{equation}
where $\textbf{Q}_n$ denotes text embedding of query $q_n$, $||$ denotes row-wise concatenation, and $\Leftrightarrow$ indicates labeled correspondence between query and video. 
Therefore, $\textbf{X}^{q}=\textbf{Q}_s$ and $\textbf{X}^{v}=\textbf{V}$ in Eq. (\ref{eq:mu}).

\textit{\textbf{Multi-granularity Aggregation:} }
Estimating Gaussian distributions requires an effective aggregator $g^m(\cdot)$ to extract holistic features. To ensure representation fidelity, we introduce multi-granularity aggregation for the sequential ${\bf V}$ and ${\bf Q}_s$, which preserves local-global contextual cues before projecting them into a probabilistic space. Specifically, we apply mean pooling and linear mapping to obtain a global feature $\mathbf{x}^{m,g}$ and gated attention \citep{Lin2017satt,vaswani2017attention} to extract fine-grained local semantics $\mathbf{x}^{m,l}$. 
Formally, 
\begin{equation}
\left\{\begin{aligned}
    \mathbf{x}^{m,g} &= \text{FC}^m(\text{MeanPool}(\mathbf{X}^m)),\\
    \mathbf{x}^{m,l} &=\text{Softmax}(\mathbf{w}_2 \text{Tanh}(\mathbf{W}_1 \mathbf{X}^m))\cdot \mathbf{X}^m, 
\end{aligned}\right. 
\end{equation}
{where $\mathbf{W}_1 \in \mathbb{R}^{d \times d}$ and $\mathbf{w}_2 \in \mathbb{R}^{d}$ are trainable parameters.}
Then, we integrate local and global information, obtaining the multi-granularity holistic representations of $\mathbf{X}^m$ as:
\begin{equation}
   g^m(\mathbf{X}^m) = \text{LayerNorm}(\mathbf{x}^{m,g}+\mathbf{x}^{m,l}).
\vspace{-0.2cm}
\end{equation}

\noindent\textbf{(2) Joint Video and Text Domain.}
After obtaining the probabilistic distributions for video and text, we use two complementary loss functions to enforce a structured joint embedding space. 

\textit{\textbf{Distribution Alignment Loss:}} To establish consistency between video and text distributions, we introduce a distribution alignment loss $\mathcal{L}_{\mathrm{DA}}$, which minimizes the Kullback-Leibler (KL) divergence between their probabilistic representations. Additionally, an auxiliary KL regularization item is used to encourage both distributions to approach a standard normal prior \( \mathcal{N}(0, I) \) \citep{wang2024decoupling}. $\mathcal{L}_{\mathrm{DA}}$ is defined as:
\begin{equation}
    \begin{aligned}
    \mathcal{L}_{\mathrm{DA}}=&  \text{KL}\left(p(\mathbf{z}^{q}|\mathbf{x}^{q})\|p(\mathbf{z}^{v}|\mathbf{x}^{v})\right)\\
    &+\sum_{m\in \{q,v\}}\text{KL}\left(p(\mathbf{z}^{m}|\mathbf{x}^{m})\|\mathcal{N}(0,\mathbf{I})\right).
    \end{aligned}
\end{equation}

\textit{\textbf{Proxy Matching Loss:} }
In PRVR, multiple semantic relationships exist between queries and untrimmed videos, making one-to-one matching insufficient. We therefore adopt a proxy matching loss $\mathcal{L}_{\mathrm{PM}}$, which considers multiple candidate alignments to enhance robustness in representation learning.
Using the reparameterization technique \citep{kingma2013auto}, we generate $K$ proxy embeddings from the learned distributions as: 
\begin{equation}
 \begin{aligned}
    \mathbf{\hat z}^m_{k}&=\bm \mu^{m}+\bm \sigma^{m}\cdot \epsilon_k,\ \ k=\{1,..,K\},
 \end{aligned}
\end{equation}
where $\epsilon^k \sim \mathcal{N}(0, \mathbf{I})$ and $\mathbf{\hat z}^m_{k}$ is the $k$-th proxy embedding for input $m$. ${\bm \mu}^m$, ${\bm \sigma}^m$ are the mean and standard deviation calculated by Eq. (\ref{eq:mu}).
This allows the model to sample diverse but semantically related embeddings, promoting the robustness of semantic alignment.
 
For each text proxy \( \mathbf{\hat z}^q_k\), the positive video set \( \mathcal{P} = \{\mathbf{\hat z}^v_{k}\}_{k=1}^{K} \) consists of $K$ video proxies from $v$, and the negative video set \( \tilde{P}= \{\mathbf{\hat z}^{\tilde v}_{k}\}_{{\tilde v},k},{\tilde v} \neq v\) includes proxies from other videos in the batch. We then employ a multi-instance InfoNCE loss \citep{miech2020end, fang2023uatvr} to maximize the similarity between positive pairs while pushing apart negatives:
\begin{align}
\mathcal{L}_{\mathrm{PM}} = -\frac{1}{|\mathcal{B}|}\sum_{(q,v) \in \mathcal{B}} \log\frac{\sum_{\mathbf{\hat z}^v_k\in\mathcal{P}} e^{\text{cos}(\mathbf{\hat z}^q_k, \mathbf{\hat z}^v_k)/\tau}}{\sum_{\mathbf{\hat z}^v_k\in\{\mathcal{P}\cup\widetilde{\mathcal{P}}\}} e^{\text{cos}(\mathbf{\hat z}^q_k, \mathbf{\hat z}^v_k)/\tau}},
\end{align}
where $\tau$ is a temperature factor and $\mathcal{B}$ is mini-batch.

\begin{table*}[t] 
    \centering
    \caption{Performance comparison. Models are sorted in ascending order in terms of SumR on TVR. 
    }
    \vspace{-0.2cm}
    \label{tab:model_performance}
    \renewcommand\arraystretch{1.05}
    \resizebox{0.95\textwidth}{!}{ \setlength{\tabcolsep}{1.5mm}{
    \begin{tabular}{lccccclccccl}
    \toprule[1pt]
    \multirow{2}{*}{Model} &\multirow{2}{*}{{Venue}} & \multicolumn{5}{c}{TVR} & \multicolumn{5}{c}{ActivityNet} \\
     && R@1 & R@5 & R@10 & R@100 & SumR & R@1 & R@5 & R@10 & R@100 & SumR \\
    \hline
    \multicolumn{11}{l}{T2VR models:} \\ \hline
    DE++ \citep{dong2021dual} & TPAMI'21 & 8.8 & 21.9 & 30.2 & 67.4 & 128.3 & 5.3 & 18.4 & 29.2 & 68.0 & 121.0 \\
    CLIP4Clip \citep{luo2022clip4clip} & ArXiv'21 & 9.9 & 24.3 & 34.3 & 72.5 & 141.0 & 5.9 & 19.3 & 30.4 & 71.6 & 127.3 \\
    Cap4Video \citep{wu2023cap4video} & CVPR'23 & 10.3 & 26.4 & 36.8 & 74.0 & 147.5 & 6.3 & 20.4 & 30.9 & 72.6 & 130.2 \\
    UMT-L \citep{li2023unmasked} & ICCV'23 & 13.7 & 32.3 & 43.7 & 83.7 & 173.4 & 6.9 & 22.6 & 35.1 & 76.2 & 140.8 \\
    InternVideo2 \citep{wang2024internvideo2}  & ECCV'24 & 13.8 & 32.9 & 44.4 & 84.2 & 175.3 & 7.5 & 23.4 & 36.1 & 76.5 & 143.5 \\ \hline
    \multicolumn{11}{l}{VCMR models w/o moment localization:} \\ \hline
    XML \citep{lei2020tvr} & ECCV'20 & 10.0 & 26.5 & 37.3 & 81.3 & 155.1 & 5.3 & 19.4 & 30.6 & 73.1 & 128.4 \\
    ReLoCLNet \citep{zhang2021video} & SIGIR'21 & 10.7 & 28.1 & 38.1 & 80.3 & 157.1 & 5.7 & 18.9 & 30.0 & 72.0 & 126.6 \\
    QCLPL\citep{zhang2025qclpl} & TCSVT'25 & 11.0 & 28.9 & 39.6 & 81.3 & 160.8 & 6.5 & 20.4 & 31.8 & 74.3 & 133.1 \\ 
    JSG \citep{chen2023JSG} & ACM MM'23  & 11.3 & 29.1 & 39.6 & 80.9 & 161.0 & 6.7 & 22.5 & 34.8 & 76.2 & 140.3 \\ \hline

    \multicolumn{11}{l}{PRVR models:} \\ \hline
    MS-SL \citep{dong2022partially}  & ACM MM'22 & 13.5 & 32.1 & 43.4 & 83.4 & 172.4 & 7.1 & 22.5 & 34.7 & 75.8 & 140.1 \\
    PEAN \citep{jiang2023progressive} & ICME'23 & 13.5 & 32.8 & 44.1 & 83.9 & 174.2 & 7.4 & 23.0 & 35.5 & 75.9 & 141.8 \\
    GMMFormer \citep{wang2023gmmformer}  & AAAI'24 & 13.9 & 33.3 & 44.5 & 84.9 & 176.6 & 8.3 & 24.9 & 36.7 & 76.1 & 146.0 \\
    BGM-Net \citep{yin2024exploiting} &TOMM'24   &14.1 &34.7 &45.9 &85.2 &179.9 &7.2 &23.8 &36.0 &76.9 &143.9 \\
    DL-DKD \citep{dong2023dual}  & ICCV'23 & 14.4 & 34.9 & 45.8 & 84.9 & 179.9 & 8.0 & 25.0 & 37.4 & 77.1 & 147.6 \\
    ARTVL \citep{cho2025ambiguity} & AAAI'25 &15.6 &36.3 &47.7 &86.3 &185.9 &8.3 &24.6 &37.4 &78.0 &148.3\\
    GMMFormer v2 \citep{wang2024gmmformer}  & ArXiv'24 & 16.2 & 37.6 & 48.8 & 86.4 & 189.1 & 8.9 & 27.1 & 40.2 & 78.7 & 154.9 \\ 
    MGAKD \citep{zhang2025multi} & TOMM'25 & 16.0 & 37.8 & 49.2 & 87.5 & 190.5 & 7.9 & 25.7 & 38.3 & 77.8 & 149.6 \\ \hline
    {MS-SL + RAL} & - & 14.5 & 34.3 & 45.8 & 84.5 & 179.1 & 7.4 & 23.4 & 35.4 & 76.7 & 143.0 \\
    {GMMFormer + RAL} & - & 15.8 & 36.4 & 47.9 & 86.0 & 186.1  & 8.4 & 25.1 & 37.2 & 77.0 & 147.7 \\
    \textbf{GMMFormer v2 + RAL} & - & \textbf{18.2} & \textbf{40.4} & \textbf{52.1} & \textbf{88.0} & \textbf{198.8} & \textbf{8.9} & \textbf{27.7} & \textbf{40.4} & \textbf{79.1} & \textbf{156.1} \\
    \toprule[1pt]
    \end{tabular}}}
    \vspace{-0.4cm}
\end{table*} 

\subsection{Confidence-aware Set-to-Set Alignment}
With query and video representations, $\textbf{V}=\{\textbf{v}_j\}_{j=1}^{N_f}$ and $\textbf{Q}=\{\textbf{q}_i\}_{i=1}^{L}$, we can obtain similarity matrix $\textbf{S} \in\mathbb{R}^{L\times N_f}$ via dot product, where each element represents similarity between the $i$-th query word and the $j$-th video frame. 
First, we capture the most relevant frame for each query word through max-pooling, and take the cosine similarity between them as the word-video similarity ${\bf }s_i$: 
\begin{equation}
    {s}_i = {\text{max}}(\cos({\bf q}_i,{\bf v}_1), \ldots, \cos({\bf q}_i, {\bf v}_{N_f})),
\end{equation}

To further obtain query-video similarity scores, existing methods often apply mean-pooling over $\{s_i\}_{i=1}^L$. However, some words (\eg, function words) can introduce noise to cross-modal alignment. To overcome these limitations, we propose to dynamically assign word-level confidence scores $\textbf{g} = \{g_i\}_{i=1}^L\in\mathbb{R}^{L}$ through a learnable predictor. By using the predicted $\textbf{g}$, we weight the similarities $s_i$  to compute the final query-video similarity:
\begin{equation}
    S(q,v) = \sum_{i = 1}^{L} g_i s_i, \ \  \textbf{g} = \text{MLP}({\bf Q}),
\end{equation}
where MLP consists of two linear layers and an activation function. 
The $S(q,v)$ is directly supervised by the basic retrieval loss $\mathcal{L}_{base}$ \citep{dong2022partially, wang2023gmmformer}. Therefore, the full model including MSRA and CSA modules is jointly end-to-end optimized by the total loss:
\begin{gather}
  \mathcal{L} = \mathcal{L}_{base}+ \lambda_{3}\mathcal{L}_{DA}+\lambda_{4}\mathcal{L}_{PM},
\label{eq:loss}
\end{gather}
where $\lambda_{3}$ and $\lambda_{4}$ are hyperparameters to balance the losses.

\section{Experiments}
\subsection{Experimental Setup}
\textbf{Datasets and Metrics.}
We adopt two large-scale video datasets, \ie, ActivityNet Captions (ActivityNet) \citep{krishna2017dense} and TV Show Retrieval (TVR) \citep{lei2020tvr}. Notably, timestamp annotations are unavailable for PRVR. \textbf{TVR} contains 21,793 videos collected from six television shows. Each video is associated with five natural language sentences describing different moments. The average video length is about 76 seconds. \textbf{ActivityNet} contains 20,000 YouTube videos, with an average duration of about 118 seconds. Each video has about 3.7 moments with corresponding sentence descriptions. We abide by the popular data partition used in \citep{dong2022partially}.

Following \citep{wang2023gmmformer,jiang2023progressive}, we use rank-based metrics to evaluate the model, namely R@$M$ ($M$ = 1, 5, 10, 100). R@$M$ measures the proportion of queries that correctly retrieve the target videos in the top $M$ results.  We also report the sum of all R@$M$ scores (SumR) for overall comparisons. All metrics are reported as percentages (\%). 

\textbf{Implementation Details.} Following existing methods \citep{dong2022partially}, we use ResNet \citep{he2016deep} and I3D \citep{carreira2017quo} for visual feature extraction and RoBERTa \citep{liu2019roberta} for text feature extraction on ActivityNet and TVR. In $\mathcal{L}_{PM}$, we set the number of proxies to $K=6$. The loss coefficients are set to $\lambda_{1}$=0.05, $\lambda_{2}$=1, $\lambda_{3}$=0.001, and $\lambda_{4}$=0.004. We use the Adam optimizer with a learning rate of 1$e$-4, a batch size of 128, and train for 100 epochs. An early stopping strategy is applied, terminating training if SumR does not improve within 10 epochs. All experiments are conducted on a single A800 GPU.
To gain insight into the effectiveness and generalization ability of our proposed approach, we integrate MSRA and CSA modules into three baselines: MS-SL \citep{dong2022partially}, GMMFormer \citep{wang2023gmmformer}, and GMMFormer-v2 \citep{wang2024gmmformer}.  
More implementation details are provided in the supplementary.

\begin{figure}[t]
    \centering
    \includegraphics[width=0.9\columnwidth]{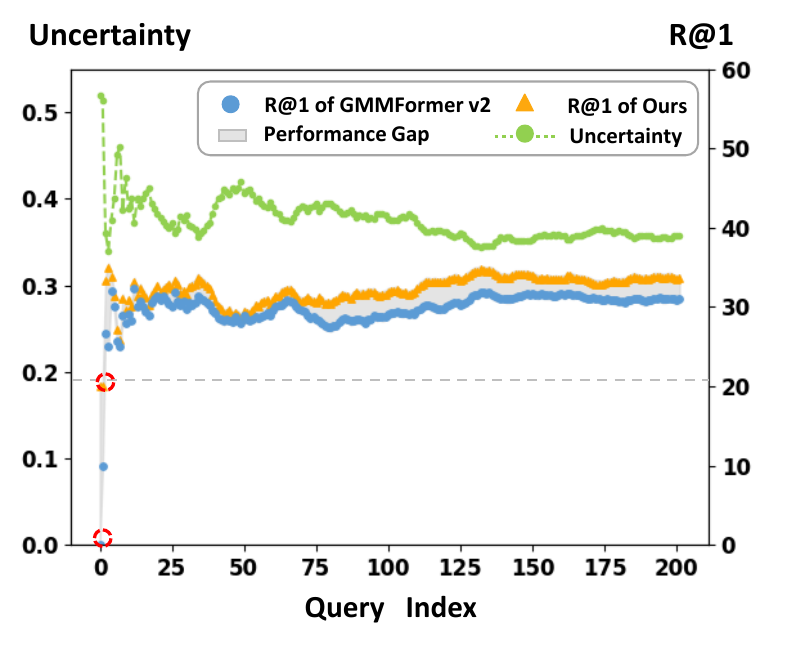}
    \vspace{-0.4cm}
    \caption{Performance comparison between our model and GMMFormer v2 under different levels of uncertainty in queries. Our model consistently outperforms GMMFormer v2, especially under extreme uncertainty.}
    \label{fig:uncert}
    \vspace{-0.4cm}
\end{figure}

\subsection{Performance Comparison}
\textbf{Effectiveness on PRVR Task.} Our method allows seamless integration into the various baseline models. As shown in Table \ref{tab:model_performance}, we apply it to three advanced PRVR models, MS-SL, GMMFormer, and GMMFormer v2. The experimental results reveal two key findings: (1) our method consistently enhances all baseline models and achieves substantial performance gains across two datasets; (2) our method sets a new state-of-the-art performance for PRVR, with a SumR of 198.8 on TVR, remarkably surpassing the previous best model (GMMFormer v2) by 9.7. These findings validate the effectiveness of our approach across the different architectures. In the following parts, we adopt GMMFormer v2 as our default benchmark model for further analysis and comparisons. 

\textbf{Model Robustness on Uncertain Samples.}
To verify the robustness and stability of our proposed method, we conduct more comparisons on queries with different uncertainty levels and observe the R@1 score. For clarity, we select a subset of TVR test set (\ie, queries with M/V ratio $\in$[0.2,0.4] \citep{dong2022partially}) and group every 5 queries into a set. 
The uncertainty level of each query set is quantified using the geometric mean \citep{gao2024embracing} of ${\bm \sigma}^q$ in Eq. (\ref{eq:mu}). 
By observing the experimental results in Figure \ref{fig:uncert}, we can find that: (1) our method consistently outperforms GMMFormer v2 across different uncertainty levels; (2) the performance gap between our method and GMMFormer v2 widens as uncertainty increases; (3) under extreme uncertainty, GMMFormer v2 collapses, with R@1 approaching zero, while our model remains stable, achieving an R@1 of nearly 20.
These findings demonstrate the effectiveness of our method in mitigating the impact of data uncertainty, ensuring robust query-video alignment even in \textit{highly ambiguous cases}.

\begin{figure}[t]
    \centering
    \includegraphics[width=\columnwidth]{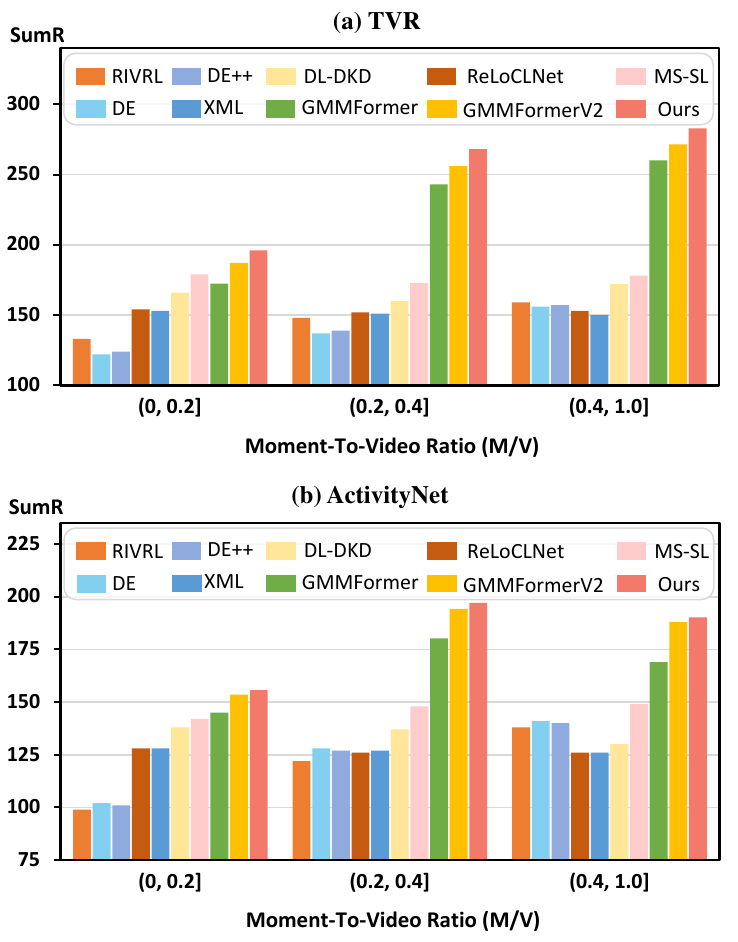}
    \caption{Performance on different types of queries. Queries are grouped according to their M/V ratio $r$. The smaller $r$ indicates less relevant content while more irrelevant content to the query.}
    \vspace{-0.5cm}
    \label{fig:ratio}
\end{figure}

\textbf{M/V Performance Analysis.}
In PRVR, queries capture only partial aspects of the video content. Here, we analyze the performance across queries with different M/V ratios $r$, namely the proportion of query-relevant moment to the total video length. A smaller $r$ indicates that the target video contains less relevant content. This semantic imbalance between the query and the video makes retrieval more challenging. Following \citep{dong2022partially}, we categorize the test queries into three groups: short ($r \in(0, 0.2]$), medium ($r \in(0.2, 0.4]$), and long ($r \in(0.4, 1.0]$). As shown in Figure \ref{fig:ratio}, our model consistently outperforms others, demonstrating its effectiveness and robustness across queries with varying relevance levels.

\subsection{Further Analysis}
\begin{figure}[t]
    \centering
    \includegraphics[width=\columnwidth]{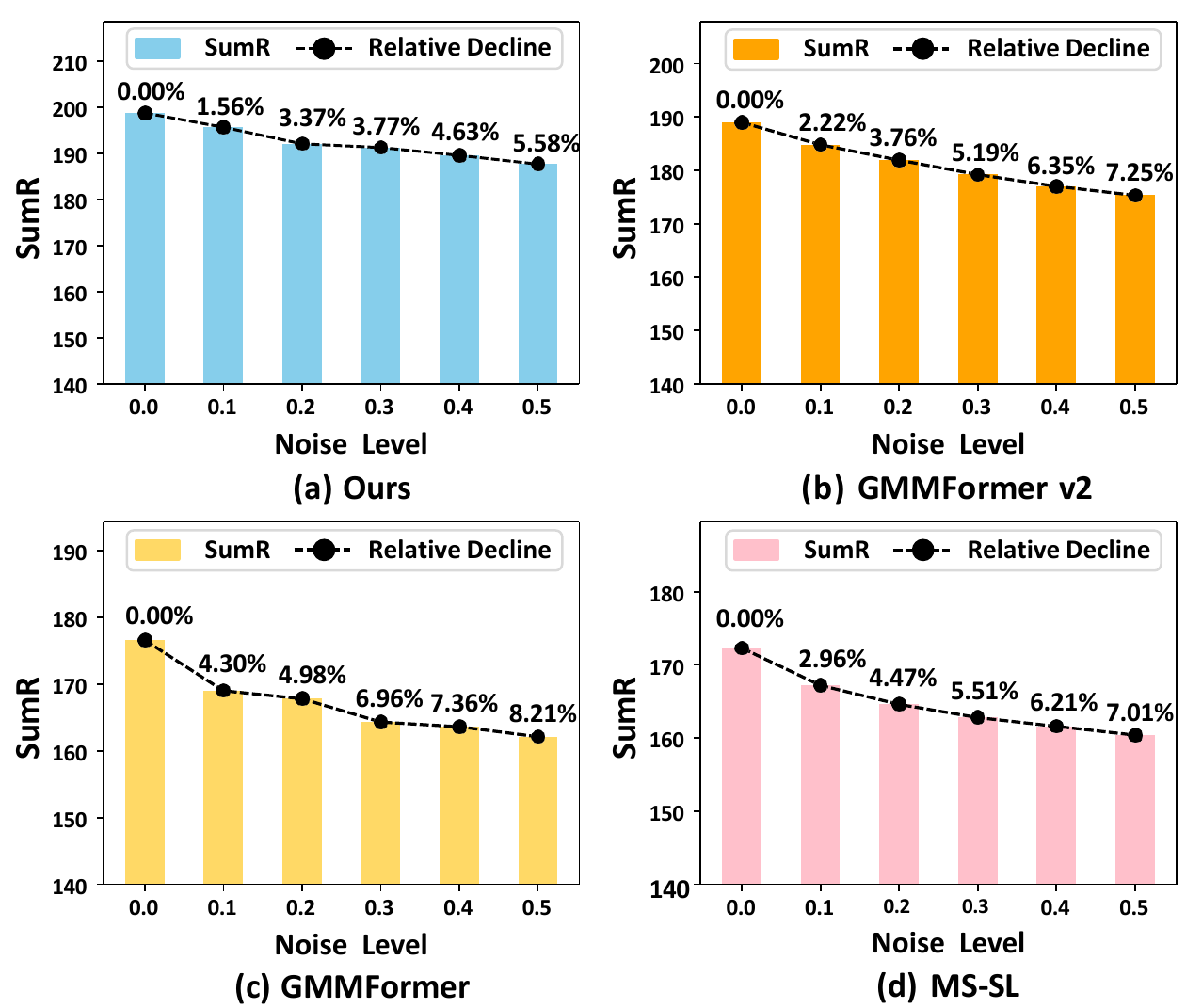}
    \caption{Performance of different methods under different levels of noise on TVR. Our model exhibits \textbf{the smallest performance drop} as noise level increases.}
    \vspace{-0.1cm}
    \label{fig:noise}
\end{figure}

\textbf{Model Robustness Under Noise.} Performance under noisy conditions poses a greater challenge to model robustness \citep{yang2024learning,pan2024finding}. Following \citep{yang2021deconfounded}, we insert a randomly generated segment with a duration of \(h \times p\) seconds at the beginning of the test video, where $h$ represents the duration of the test video and \(p\) denotes the noise level. As shown in Figure \ref{fig:noise}, our model consistently outperforms comparison methods under different noise levels and exhibits the smallest performance drop as noise intensity increases. This highlights the superior resilience of our uncertainty-aware alignment strategy to noisy inputs.

\begin{table}[t]
    \centering
    \caption{Ablation studies on model structure on TVR. 
    } 
    \label{tab:ab_comp} 
    \vspace{-0.2cm}
    \renewcommand\arraystretch{1.2}
    \resizebox{0.9\columnwidth}{!}{ \begin{tabular}{cccccccc}
        \toprule
        MSRA & CSA  & R@1 & R@5 & R@10 & R@100 & SumR &$\Delta$SumR \\
        \midrule
        &  &16.2 & 37.6 & 48.8 & 86.4 & 189.1 & -\\
        \checkmark & & 17.5 & 39.2 & 50.7 & 87.4 & 194.8 &\textcolor{green!70!black}{+5.7}   \\
        & \checkmark  &17.0 &38.5 &51.0 &88.1 &194.5 &\textcolor{green!70!black}{+5.4}\\
        \checkmark & \checkmark & \textbf{18.2} & \textbf{40.4} & \textbf{52.1} & \textbf{88.0} & \textbf{198.8} &\textcolor{green!70!black}{+9.7}\\
        \bottomrule
    \end{tabular}}
    \vspace{-0.4cm}
\end{table}

\textbf{Analysis on Model Structure.}
We provide an ablation study on TVR in terms of uncertain learning (\textit{w.t.f.} MSRA) and confidence-aware alignment (\textit{w.t.f.} CSA) in Table \ref{tab:ab_comp}. Firstly, we show the baseline GMMFormer v2 (top row). Based on it, we introduce the MSRA module (2nd row), obtaining 5.7 boost at R@1. This shows the superiority of introducing multimodal learning on distributional representations over plain semantic features. 
We also evaluate the effect of the CSA module (3rd row). By comparison, considering the word-level confidence of the query significantly improves the performance. This is because meaningless words in the query can capture unrelated background frames, misleading retrieval. 
By jointly using the designed MSRA and CSA, our method acquires an improvement of 9.7 on SumR (4th row). These ablations demonstrate the effectiveness of each component of our method in improving PRVR baselines.

\begin{figure}[t]
    \centering
    \includegraphics[width=0.9\columnwidth]{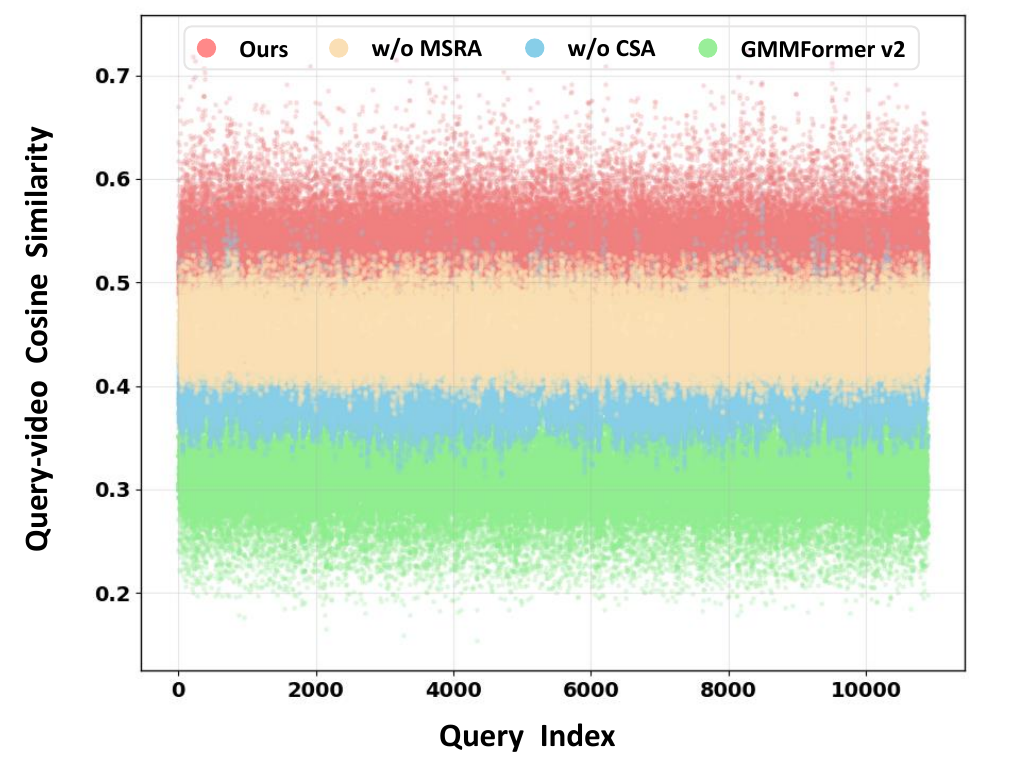}
    \vspace{-0.2cm}
    \caption{Query-video cosine similarities between test queries on TVR and their top-10 retrieved videos .} 
    \label{fig:score}
    \vspace{-0.4cm}
\end{figure}

In Figure \ref{fig:score}, we show the cosine similarities between queries on TVR and the top 10 retrieved videos given by different models.
By and large, our model produces similarities above 0.5, while other models range from 0.2 to 0.5. 
Our complete model not only demonstrates superior retrieval performance but also retrieves videos with higher similarities, indicating that the model can achieve more stable and confident query-video alignment.

\textbf{Analysis on Distribution Optimization.}
In Table \ref{k_number}, we conduct ablation studies concerning the learning objectives on the MSRA module. 
$\mathcal{L}_{DA}$ minimizes the KL distance between the distributions of each query-video pair. $\mathcal{L}_{PM}$ promotes the semantic similarity of random video and text proxies in a contrastive learning framework. 
By observing Table \ref{k_number}, we have drawn the following conclusions: 
(1) although removing any loss term leads to a performance decline, both variants are still superior to the baseline without distribution optimization. 
(2) the joint usage of $\mathcal{L}_{DA}$ and $\mathcal{L}_{PM}$ achieves the best performance, showing the complementarity and effectiveness of constraints on the multimodal distributions and random proxies. 

We further discuss the number of sampling proxies for $\mathcal{L}_{PM}$ in Table \ref{k_number}. For ``w/o sampling'', we directly use the mean of Gaussian distribution as a proxy during the training. This gives a sub-optimal performance due to fixing features rather than exploiting data uncertainty. 
As the number of proxies $K$ increases from 2 to 6, our method enables better performance by gradually augmenting the data representation based on uncertainty. 
Considering the trade-off between the performance and computational cost, we choose $K$=6 in our final model. 

\begin{table}[t]
    \centering
    \caption{Ablation studies on distribution optimization and proxy number on TVR.} 
    \label{k_number}
    \vspace{-0.1cm}
    \renewcommand\arraystretch{1.05}
    \resizebox{0.9\columnwidth}{!}{\begin{tabular}{cccccc}
    \toprule[1pt]
    {Loss}   & R@1  & R@5       & R@10   & R@100  & SumR \\  \hline
    \textit{w/o} $\mathcal{L}_{DA}$ & 17.7 & 39.9 & 51.8 & 88.0 & 197.4 \\  
    \textit{w/o} $\mathcal{L}_{PM}$ & 17.4 & 39.7 & 51.6 & 87.8 & 196.5 \\   \hline\hline
    Proxy   & R1  & R5       & R10   & R100  & SumR \\  \hline
    \textit{w/o} sampling   & 17.6  & 39.8       & 51.7   & 87.6  & 196.7 \\ 
    
    $K$=2  & 17.9  & 40.1  & 51.8   & 87.7  & 197.5  \\
    $K$=4   & 18.0   & 40.3   & 51.8 & 87.9  & 198.0  \\
    $K$=\textbf{6} & \textbf{18.2} & 40.4 & \textbf{52.1} & \textbf{88.0} & \textbf{198.8} \\
    \toprule[1pt]
    \end{tabular}}
    \vspace{-0.1cm}
\end{table}

\begin{table}[t]
    \centering
    \caption{Effect of different uncertainty modeling methods on TVR.}
    \label{dim_gauss}
    \vspace{-0.2cm}
    \renewcommand\arraystretch{1.15}
    \resizebox{0.9\columnwidth}{!}{ \begin{tabular}{lccccc}
    \toprule[1pt]
    \multirow{1}{*}{Method} & R@1 & R@5 & R@10 & R@100 & SumR \\
    \hline
    $\textbf{X}^{q}=\textbf{Q}$ & 17.6 & 39.8 & 51.5 & 87.9 & 196.9 \\
    $g^m(\mathbf{X}^m) =\mathbf{x}^{m,l}$  & 17.9 & 40.2 & 51.2 & 87.4 & 196.7 \\
    $g^m(\mathbf{X}^m) = \mathbf{x}^{m,g}$  & 18.0 & 40.1 & 51.6 & 88.2 & 197.9 \\ 
    \textbf{Ours} & \textbf{18.2} & \textbf{40.4} & \textbf{52.1} & \textbf{88.0} & \textbf{198.8} \\
    \toprule[1pt]
    \end{tabular}}
    \vspace{-0.3cm}
\end{table}

\begin{figure*}[t]
  \centering
  \includegraphics[width=\textwidth]{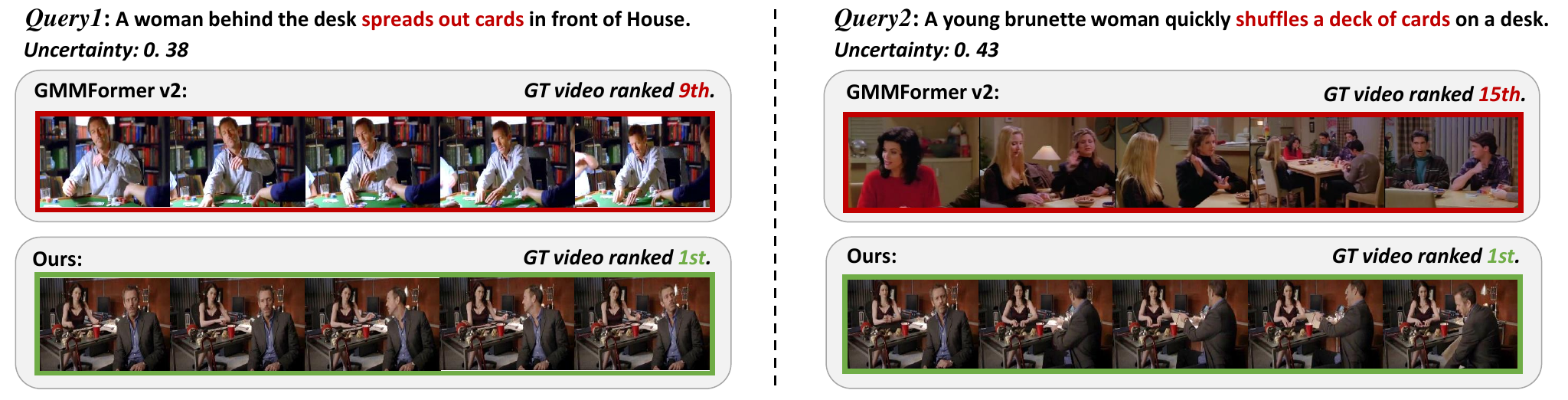}
  \vspace{-0.6cm}
  \caption{PRVR results on TVR: top-1 retrieved video by our method and GMMFormer v2 \citep{wang2024gmmformer}. \textcolor{Mygreen}{Green} and \textcolor{Myred}{red} boxes indicate the {ground truth} and {distractor videos}, respectively.
  }
  \label{fig:visual}
  \vspace{-0.3cm}
\end{figure*}

\textbf{Analysis on Uncertainty Modeling.} 
Table \ref{dim_gauss} highlights the effect of our key design choices in uncertainty modeling.
First, we examine the impact of our query support sets. By reducing the query support set to a single query, the text distribution fails to capture broader contextual semantics. This results in a severe semantic mismatch between the text and video distributions, disrupting the optimization process and significantly degrading performance. 
Next, we explore the role of multi-granular feature aggregation in quantifying data uncertainty. The global aggregation summarizes the holistic context while local aggregation supplements fine-grained details. The results reveal that the combination of global-local aggregation contributes to robust uncertainty modeling and achieves the best performance.

\begin{figure}[t]
    \centering
    \includegraphics[width=\columnwidth]{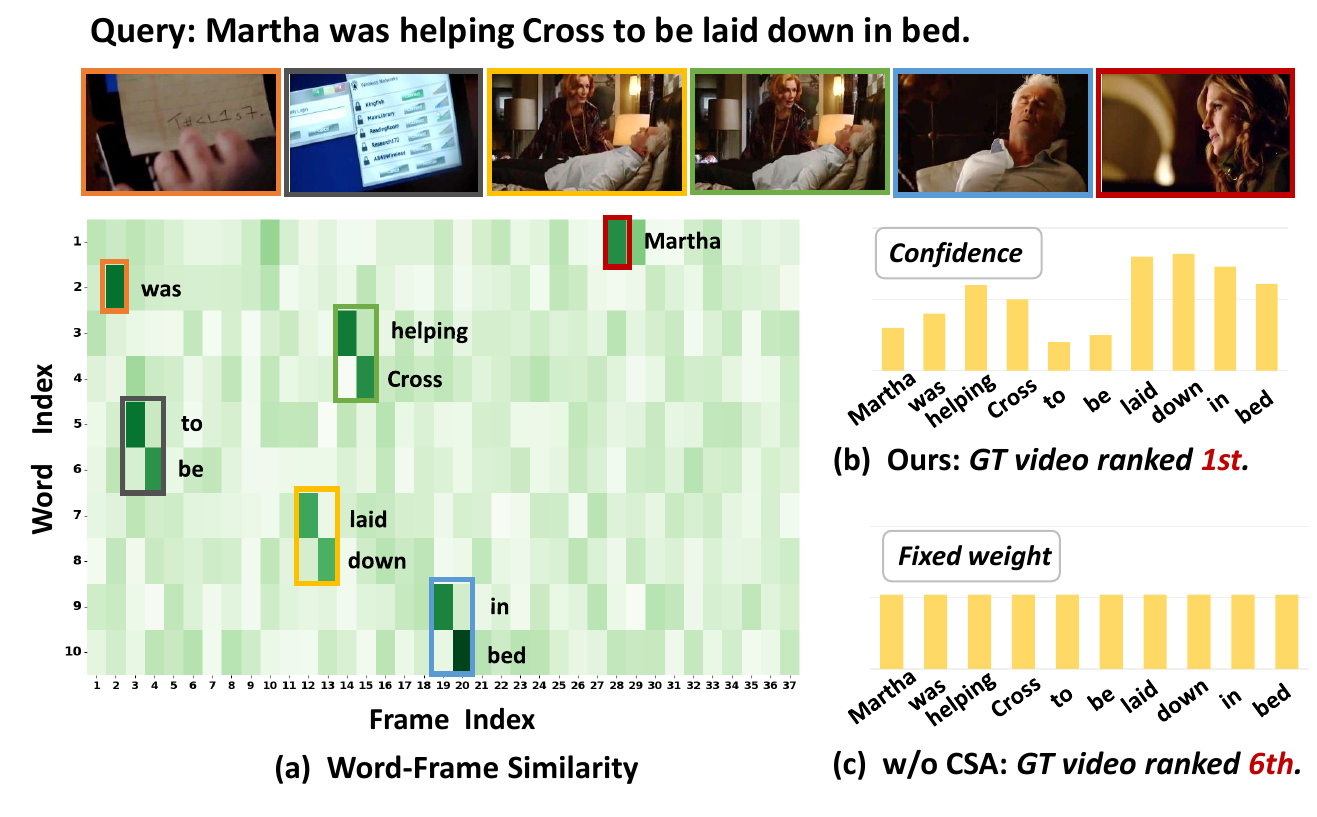}
    \vspace{-0.6cm}
    \caption{Visualization of CSA mechanism. (a) Word-frame similarity shows uninformative words (\eg, ``to be'') align with irrelevance frames (gray box). (b) With CSA, uninformative words receive lower confidence, improving retrieval. (c) Fixed average weight causes performance drop, with GT video ranked 6th. 
    }
    \label{fig:confiden}
    \vspace{-0.4cm}
\end{figure}

\subsection{Qualitative Results}
\textbf{Challenging Retrieval Cases.} 
To further investigate the impact of data uncertainty, we analyze two challenging retrieval cases with data uncertainty. As shown in Figure \ref{fig:visual}, we display two queries that refer to different moments within the ground-truth (GT) video and compare the Top-1 retrieval results of our model and GMMFormer v2. 
We find GMMFormer v2 fails in both cases, retrieving distractor videos containing similar actions (\textit{spreading out cards} or \textit{shuffling a deck of cards}), while ranking GT video at position 9th and 15th. In contrast, our model effectively excavates semantic relationships in data and retrieves GT video at Rank-1.

\textbf{Confidence-aware Alignment.} 
Here, we investigate how our proposed CSA improves retrieval performance. Figure \ref{fig:confiden}a shows the word-frame similarity matrix, where words like ``to be'' are aligned with frames of low relevance to the query (highlighted in gray box). By introducing CSA (Figure \ref{fig:confiden}b), the words ``to be'' receive a lower confidence, resulting in the correct retrieval of the GT video. In contrast, when confidence is replaced with a fixed average weight (Figure \ref{fig:confiden}c), the GT video drops to the 6th rank, demonstrating the importance of dynamical confidence weighting for precise retrieval.

\subsection{Versatility on T2VR} 
T2VR can be viewed as a simple case of PRVR, where videos are trimmed to correspond to queries. In Table \ref{tab:t2vr}, we apply our RAL to T2VR task, comparing it with CLIP4Clip \citep{luo2022clip4clip} under two different visual backbones. It can be found that combining RAL with CLIP4Clip improves the R@1 by about 6.1\% and 6.6\% under ViT-B/32 and ViT-B/32, respectively.
The results further demonstrate the effectiveness and versatility of our framework in enhancing cross-modal alignment.

\begin{table}[t]
\centering
\caption{Text-to-video performance of our method on MSR-VTT dataset for the T2VR task.}
    \vspace{-0.2cm}
    \label{tab:t2vr}
    \renewcommand\arraystretch{1.05}
    \setlength{\tabcolsep}{4.1pt}
\resizebox{\columnwidth}{!}{ \begin{tabular}{l|ccccc}
\toprule[1pt]
          {Method} &
  \multicolumn{1}{c}{R@1} &
  \multicolumn{1}{c}{R@5} &
  \multicolumn{1}{c}{R@10} &
  \multicolumn{1}{c}{$\text{MdR} {\textcolor{green!70!black} \downarrow}$} &
  \multicolumn{1}{c}{$\text{MnR} {\textcolor{green!70!black} \downarrow}$} \\ \hline
CLIP4Clip (ViT-B/32)  & 44.5  & 71.4 & 81.6 & 2.0 & 15.3 \\
\textbf{+RAL} & \textbf{47.2} &  \textbf{73.6} & \textbf{83.1} & \textbf{2.0} &  \textbf{12.5}  \\ \hline
CLIP4Clip (ViT-B/16)  & 47.1  & 74.1 & 81.8 & 2.0 & 14.9 \\
\textbf{+RAL}     & \textbf{50.2}  & \textbf{76.1} & \textbf{85.2} & \textbf{1.0} & \textbf{12.7} \\
\toprule[1pt]
\end{tabular}
}
\vspace{-0.4cm}
\end{table}

\section{Conclusion}
\vspace{-0.2cm}
In this paper, we investigate the fundamental challenge of spurious semantic correlations, which arises from query ambiguity and partial video relevance. 
We propose a novel Robust Alignment Learning (RAL) framework that explicitly models data uncertainty by representing both video and text features as probabilistic distributions, enabling more robust cross-modal alignment. We introduce a query support set that aggregates multiple descriptions of the same video, and multi-granularity feature aggregation to quantify data uncertainty more effectively. Additionally, we design a confidence-aware set-to-set alignment mechanism to assign adaptive weights to query words, improving retrieval precision. 
Extensive experiments on benchmark datasets demonstrate the effectiveness and versatility of our RAL, achieving significant improvements in both PRVR and T2VR.

\section*{Limitations}
In the validation experiments on the TVR dataset, we conducted an attribution analysis of retrieval failure cases and identified a prominent pattern: \textit{cross-modal alignment bias caused by missing named entities}. For instance, in the query ``Beckett confronts a friend at the bar", the discrepancy between the model’s retrieved result and the ground-truth (GT) video stems from the model's failure to associate the textual character entity ``Beckett" with the corresponding visual representation in the video. Specifically, the GT video contains distinctive visual cues associated with this character—such as a red jacket and curly hair. In contrast, the retrieved distractor video, although set in a similar bar scene, lacks these fine-grained identity indicators. Our current approach does not explicitly model the correspondence between named entities in the query and specific characters in the video, leading to retrieval ambiguity. This limitation highlights a potential direction for future research: \textit{incorporating identity-aware modeling} to associate textual mentions of characters with their visual counterparts in the video \citep{song2024efficiently,zhou2024scene,zhou2025egotextvqa}. This could involve integrating entities attribute information from knowledge graphs and using attention mechanisms to guide the model toward identity-relevant visual cues, thereby enhancing its applicability in real-world retrieval scenarios \citep{zhang2024visual,song2024emotional}.

\section*{Acknowledgements}
This work was supported in part by the National Natural Science Foundation of China under Grant 62402471, Grant U22A2094, Grant 62472385, and Grant 62272435. We also acknowledge support from the Pioneer and Leading Goose R\&D Program of Zhejiang under Grant 2024C01110. This research was also supported by the advanced computing resources provided by the Supercomputing Center of the USTC. We also acknowledge the support of GPU cluster built by MCC Lab of Information Science and Technology Institution, USTC.


\bibliography{custom}

\appendix

\section*{Example Appendix}
\label{sec:appendix}
This supplementary document includes the following:
\vspace{-2mm}
\begin{itemize}
\setlength{\itemsep}{0pt}
\setlength{\parsep}{0pt}
\setlength{\parskip}{0pt}
    \item[(\textit{i})] Additional information about our implementation details (Section \ref{Imple_Details});
    \item[(\textit{ii})] Additional experimental results and analysis (Section \ref{Exp:all}), including the variation trend of uncertainty and performance during training (Section \ref{Exp:uncert}), the impact of loss coefficients (Section \ref{Exp:hy}), and studies on the retrieval efficiency of different PRVR methods (Section \ref{Exp:efficiency});
    {\item[(\textit{iii})] Additional qualitative examples of our method and discussions on future work (Section \ref{visual}).}
\end{itemize}

\begin{figure}[h]
    \centering
    \includegraphics[width=\columnwidth]{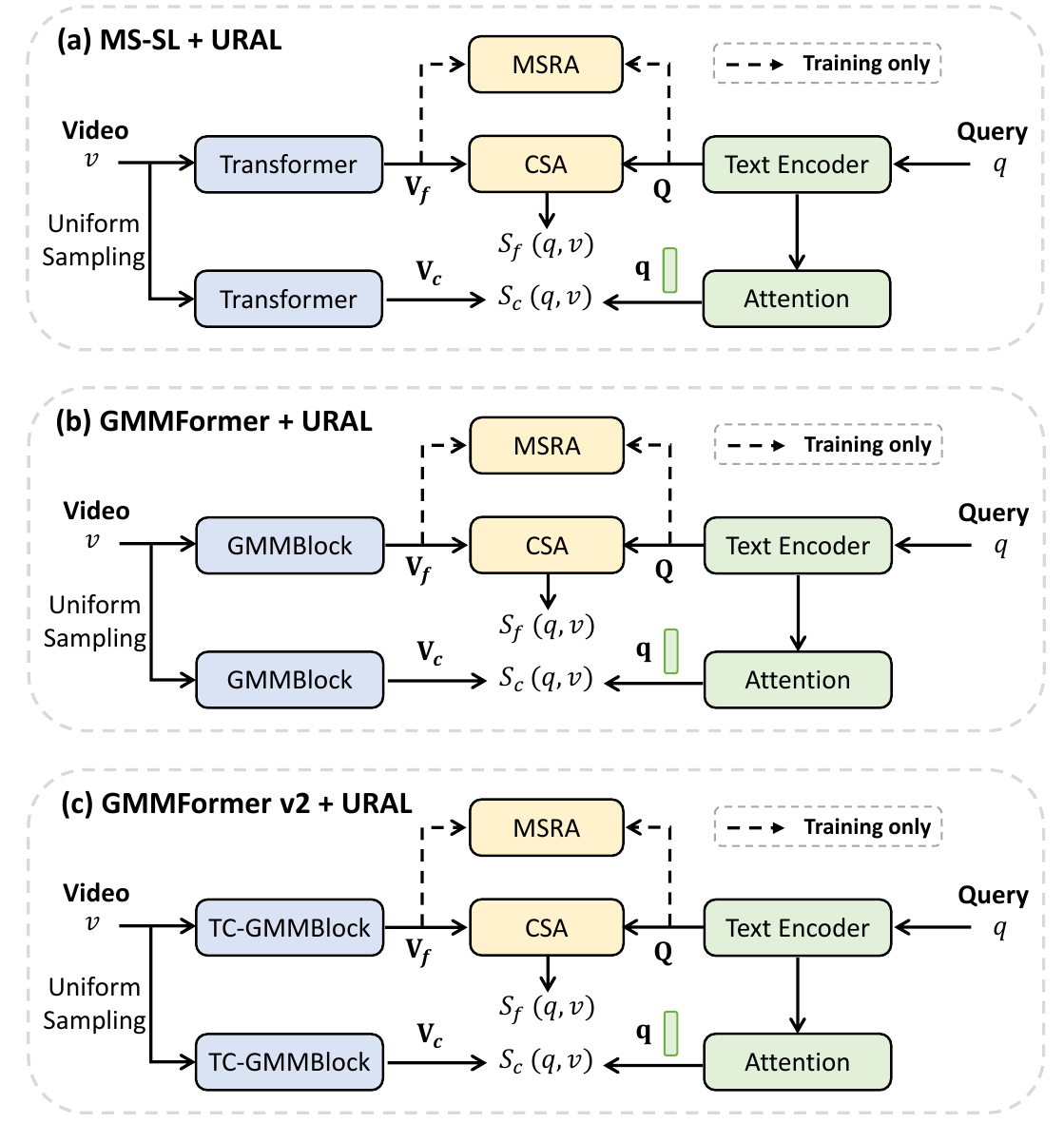}
    \caption{Integration of RAL with PRVR baselines, where MSRA and CSA stand for the proposed multimodal semantic robust alignment and confidence-aware set-to-set alignment modules, respectively. RAL is integrated into the frame-level branch, and the final retrieval score is a combination of frame-level and clip-level scores.}
    \label{fig:method}
    \vspace{-0.4cm}
\end{figure}

\section{Implementation Details}
\label{Imple_Details}
Figure \ref{fig:method} elaborates on the experimental details for applying the proposed URAL framework on existing baselines, including MS-SL \citep{dong2022partially}, GMMFormer \citep{wang2023gmmformer}, and GMMFormer v2 \citep{wang2024gmmformer}. Specifically, URAL is incorporated solely into the frame-level branch of the baselines. {This design choice is motivated by considerations of: (1) the frame-level branch provides fine-grained temporal information, which is essential for handling the uncertainty inherent in cross-modal alignment while avoiding unnecessary computational overhead; (2) frame-level features are more conducive to integrating our proposed modules, such as the confidence-aware set-to-set alignment module.} In our implementation, we retain the original video and text encoders from the baselines to ensure fair comparison. The extracted video frame features ${\bf V}_f$ and query word features ${\bf Q}$ are fed into the MSRA (Multimodal Semantic Robust Alignment) module, which explicitly models and mitigates uncertainty to enhance cross-modal alignment. The resulting robust ${\bf Q}$ and ${\bf V}_f$ are subsequently processed by CSA (Confidence-aware Set-to-set Alignment) module for query-video alignment with adaptive confidence weighting. The CSA module generates reliable frame-level retrieval scores $S_f(q,v)$, which are then summed with the clip-level scores $S_c(q,v)$ to produce the final retrieval result.

\section{More Experimental Results}
\label{Exp:all}

\begin{figure}[t]
    \centering
    \includegraphics[width=\columnwidth]{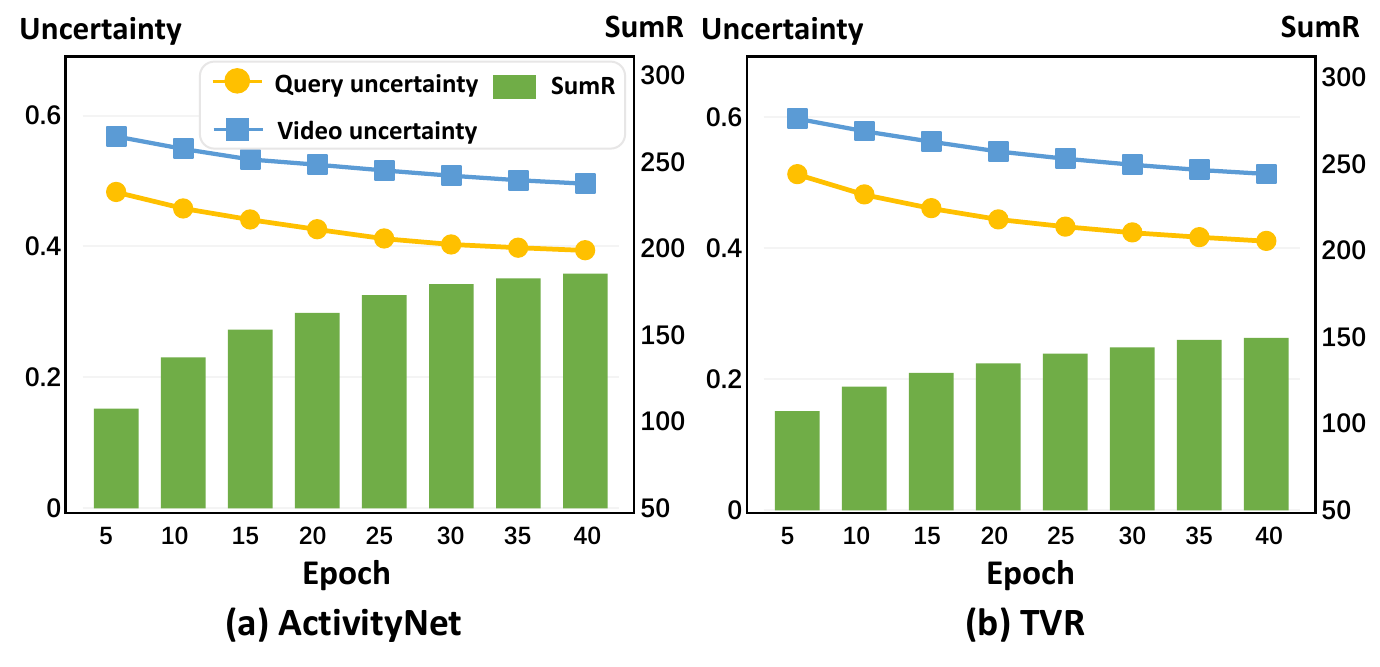}
    \caption{
    The variation trend of data uncertainty and retrieval performance during training. As training progresses, reduced uncertainty leads to improved retrieval accuracy. }
    \label{fig:epoch}
    \vspace{-0.2cm}
\end{figure}

\begin{figure}[t]
    \centering
    \includegraphics[width=0.9\columnwidth]{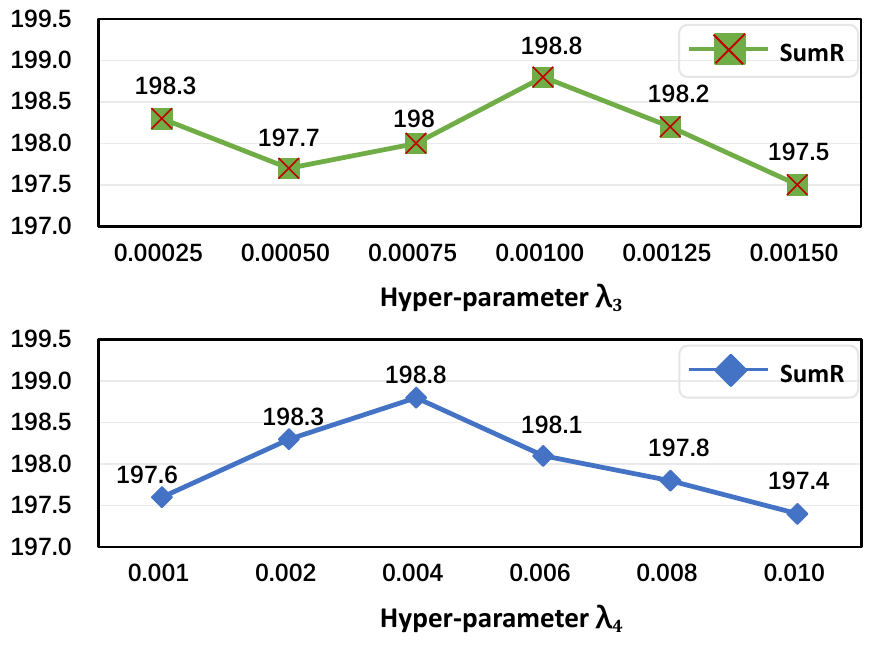}
    \caption{The impact of the loss coefficients, $\lambda_{3}$ and $\lambda_{4}$, of distribution alignment loss $\mathcal{L}_{DA}$ and proxy matching loss $\mathcal{L}_{PM}$.}
    \label{fig:lambda}
    \vspace{-0.6cm}
\end{figure}

\subsection{Uncertainty Mitigation During Training}
\label{Exp:uncert}
To investigate the variation in video and query uncertainty during training and its impact on performance, we quantify the uncertainty every 5 training epochs and report the corresponding SumR performance on the test set. We observe that the uncertainty of both videos and queries decreases as training progresses, and the model's retrieval performance improves. This indicates that mitigating uncertainty is crucial for improving retrieval accuracy. Additionally, we find that videos exhibit higher uncertainty than queries. The presence of redundant content in untrimmed videos is a major challenge for PRVR. This highlights an important direction for future research.

\subsection{Hyper-parameter Analysis} 
\label{Exp:hy}
In addition to the basic retrieval loss $\mathcal{L}_{base}$ \citep{dong2022partially}, our model incorporates auxiliary distribution alignment loss $\mathcal{L}_{DA}$ and proxy matching loss $\mathcal{L}_{PM}$ to enhance alignment. 
In Figure \ref{fig:lambda}, we study the sensitivity of two loss coefficients, $\lambda_{3}$ and $\lambda_{4}$, on the TVR dataset. The initial settings are $\lambda_{3} = 0.001$ and $\lambda_{4}= 0.00025$ to keep each loss item at the same magnitude. We adjust these hyper-parameters within a certain range to assess their impact. As shown in Figure \ref{fig:lambda}, our model maintains stable performance and reaches the optimal balance at $\lambda_{3} = 0.004$ and $\lambda_{4} = 0.001$. 

\begin{table}[t]
\centering
\caption{Comparison in terms of FLOPs (G) and parameters (M). $\Delta$ denotes our relative changes over the baseline (GMMFormer v2) for different metrics. }
\label{flop_compare}
\vspace{-0.2cm}
\resizebox{\columnwidth}{!}{ \setlength{\tabcolsep}{1.5mm}{\begin{tabular}{lccccc}
\toprule[1pt]
&MS-SL &GMMFormer &GMMFormer v2 &Ours &$\Delta$\\ 
\midrule 
FLOPs  &1.29 &1.95 &5.43 &5.75 & \textcolor{red!90!black}{+0.32}\\
Params &4.85 &12.85 &32.27 &35.53 &\textcolor{red!90!black}{+3.26}  \\ 
SumR &172.4 &176.6 &189.1 &198.8 &\textcolor{green!70!black}{+9.7} \\
\bottomrule[1pt] 
\end{tabular}}}
\end{table}

\begin{table}[t]
    \centering
    \caption{Comparisons in terms of runtime (ms) of PRVR models.}
    \vspace{-0.2cm}
    \label{tab:retrieval_efficiency}
    \setlength{\tabcolsep}{4.5pt}
    \renewcommand\arraystretch{1.15}
\resizebox{0.9\columnwidth}{!}{     \begin{tabular}{lccccc}
        \toprule[1pt]
        Database Size & 500 & 1,000 & 1,500 & 2,000 & 2,500 \\ \hline
        MS-SL \citep{dong2022partially} & 4.89 & 6.11 & 8.06 & 10.42 & 12.93 \\
        GMMFormer \citep{wang2023gmmformer} & 2.68 & 2.93 & 3.40 & 3.94 & 4.56 \\
        GMMFormer v2 \citep{wang2024gmmformer} & 3.95 & 4.32 & 5.02 & 5.81 & 6.73 \\
        Ours & 4.61 & 5.05 & 5.86 & 6.79 & 7.86 \\
        \toprule[1pt]
    \end{tabular}}
    \vspace{-0.4cm}
\end{table}

\begin{figure*}[t]
  \centering
  \includegraphics[width=\textwidth]{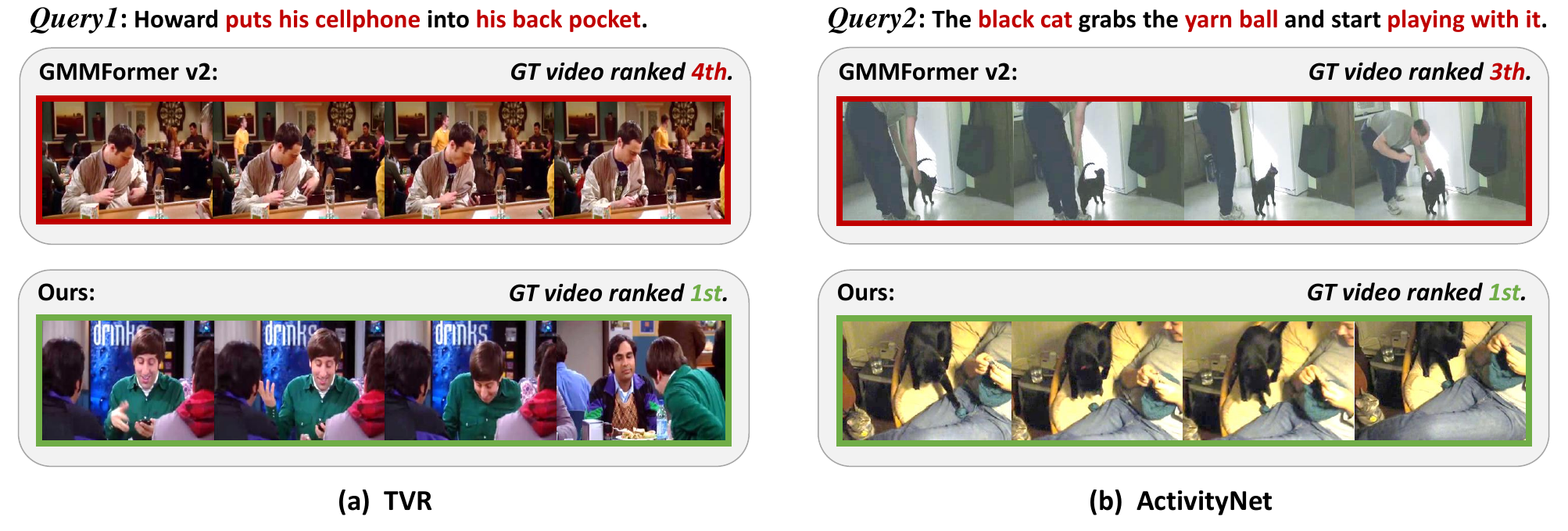}
  \caption{More visualization results on TVR and ActivityNet. Top-1 retrieved videos from our method and GMMFormer v2 \citep{wang2024gmmformer} are shown. \textcolor{Mygreen}{Green} and \textcolor{Myred}{red} boxes indicate the {ground truth} and {distractor videos}, respectively.
  }
  \label{fig:suc-visual}
  \vspace{-0.4cm}
\end{figure*}

\begin{figure}[t]
  \centering
  \includegraphics[width=\columnwidth]{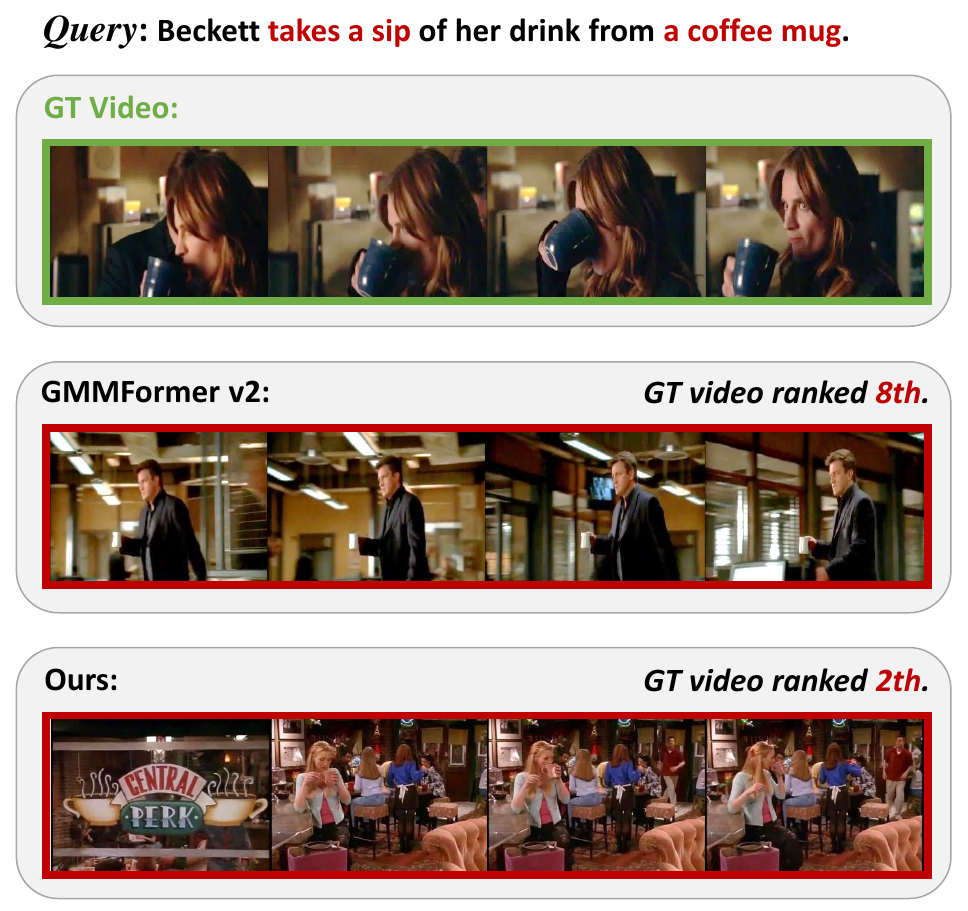}
  \caption{Failure case on TVR. \textcolor{Myred}{red} boxes indicate the top-1 retrieved video by our method and GMMFormer v2 \citep{wang2024gmmformer}. \textcolor{Mygreen}{Green} box indicates the ground truth video.
  }
  \label{fig:fail-visual}
  \vspace{-0.4cm}
\end{figure}

\subsection{Retrieval Efficiency} 
\label{Exp:efficiency}
To evaluate model efficiency, we compare several PRVR methods in terms of floating-point operations (FLOPs) and model parameters. Our method builds upon GMMFormer v2 as the baseline while introducing uncertainty learning and confidence-aware alignment. As shown in Table \ref{flop_compare}, {while our model increases FLOPs by 0.32G and parameters by 3.26M, it achieves a substantial 9.7\% improvement in SumR.} This highlights a favorable trade-off between computational cost and performance gain.

For retrieval efficiency in practical situations, we measure the retrieval speed (in milliseconds) as shown in Table \ref{tab:retrieval_efficiency}. Specifically, we construct a video subset from the TVR dataset and measure the average runtime to complete the retrieval process for a single text query under different database size settings. Despite introducing confidence-aware alignment during retrieval, our model's runtime remains comparable to GMMFormer v2. Moreover, as the database size increases, the runtime increases marginally, demonstrating the potential for large-scale applications.

\section{More Visualization Results}
\label{visual}
\subsection{Qualitative Retrieval Results}
Figure \ref{fig:suc-visual} presents two additional visualization examples from TVR \citep{lei2020tvr} and ActivityNet \citep{krishna2017dense} datasets, comparing the top-1 retrieval results of our model against GMMFormer v2 \citep{wang2024gmmformer}. In both cases, GMMFormer v2 fails to retrieve the target videos, instead selecting distractor videos with similar scenes, such as a ``\textit{cellphone}'' and a ``\textit{black cat}'', 
while ranking the ground-truth (GT) videos at the 4th and 3rd positions, respectively. In contrast, our model effectively uncovers semantic relationships and successfully ranks the GT videos at 1st. 
For example, in Figure \ref{fig:suc-visual} (a), our model is sensitive to the action ``\textit{puts his cellphone}'', whereas GMMFormer v2 retrieves a distractor video featuring a different action, ``\textit{pull out phone}". In Figure \ref{fig:suc-visual} (b), the ``yarn ball'' is a subtle but crucial visual cue that GMMFormer v2 overlooks, whereas our model successfully detects it for accurate retrieval.
These qualitative results demonstrate that our approach significantly enhances retrieval accuracy by capturing critical semantic details in query and video.

\subsection{Failure Cases and Future Work}
Figure \ref{fig:fail-visual} presents a failure case from the TVR dataset, comparing the top-1 retrieval results of our model and GMMFormer v2. The query describes a common scenario of drinking coffee. 
Although both models fail to retrieve the GT video as the top-1 result, our model correctly captures the key phrase in the query (\ie, ``\textit{takes a sip of her drink}") and retrieves a highly relevant video, ranking the GT video in 2nd place.
In contrast, GMMFormer v2 retrieves a video of ``\textit{a man carrying a coffee cup}'' and ranks the GT video only at 8th place. 

Further analysis reveals that a critical factor distinguishing the GT video from our retrieved video is the presence of ``\textit{Beckett}", a named entity in the query. Our approach does not involve the correspondence between named entities in the query and specific individuals in the video, leading to retrieval ambiguity. This limitation highlights a potential direction for future research: incorporating identity-aware modeling to associate textual mentions of people with their visual counterparts in videos, making it better suited for real-world retrieval scenarios.

\end{document}